\documentclass[twoside]{article}
\usepackage[T1]{fontenc}
\usepackage{array}
\usepackage{url}
\usepackage{verbatim}
\def\BibTeX{{\rm B\kern-.05em{\sc i\kern-.025em b}\kern-.08em
    T\kern-.1667em\lower.7ex\hbox{E}\kern-.125emX}}
\usepackage{balance}
\usepackage{amsmath,amssymb,amsfonts}
\usepackage{graphicx}
\usepackage{textcomp}
\usepackage{xcolor}
\usepackage{stfloats}
\usepackage{enumerate}
\usepackage{comment}
\usepackage{hyperref}
\usepackage{subfigure}
\usepackage{verbatim}
\usepackage{placeins}
\usepackage{algorithm}
\usepackage{algpseudocode}
\usepackage{multirow}
\usepackage{tabularx}
\usepackage{parskip}
\usepackage{tcolorbox}
\usepackage{aligned-overset}
\usepackage{svg}
\usepackage{layouts}
\usepackage{cite}
\usepackage{authblk}
\usepackage[margin=4cm]{geometry}
\usepackage{booktabs}
\usepackage{fancyhdr}
\newcommand{\LM}[1]{\textcolor{black}{{#1}}}

\begin{document}

\title{HOLMES: HOLonym-MEronym based Semantic inspection for Convolutional  Image Classifiers}

\author[1,2]{Francesco Dibitonto }
\author[1]{Fabio Garcea}
\author[3]{André Panisson}
\author[3]{Alan Perotti}
\author[1]{Lia Morra}
\affil[1]{Politecnico di Torino, Italy}
\affil[2]{EVS Embedded Vision Systems, Italy}
\affil[2]{CENTAI Institute, Turin, Italy}

\date{}

\maketitle

\begin{abstract}
Convolutional Neural Networks (CNNs) are nowadays the model of choice in Computer Vision, thanks to their ability to automatize the feature extraction process in visual tasks.  However, the knowledge acquired during training is fully sub-symbolic, and hence difficult to understand and explain to end users. In this paper, we propose a new technique called HOLMES (HOLonym-MEronym based Semantic inspection) that decomposes a label into a set of related concepts, and provides component-level explanations for an image classification model. Specifically, HOLMES leverages ontologies, web scraping and transfer learning to automatically construct {\em meronym} (parts)-based detectors for a given {\em holonym} (class). Then, it produces heatmaps at the meronym level and finally, by probing the holonym CNN with occluded images, it highlights the importance of each part on the classification output. Compared to state-of-the-art saliency methods, HOLMES takes a step further and provides information about both \textit{where} and \textit{what} the holonym CNN is looking at. It achieves so without relying on densely annotated datasets and without forcing concepts to be associated to single computational units. Extensive experimental evaluation on different categories of objects (animals, tools and vehicles) shows the feasibility of our approach. On average, HOLMES explanations include at least two meronyms, and the ablation of a single meronym roughly halves the holonym model confidence. \LM{The resulting heatmaps were quantitatively evaluated using the deletion/insertion/preservation curves. All metrics were comparable to those achieved by GradCAM, while offering the advantage of further decomposing the heatmap in human-understandable concepts. In addition, results were largely above chance level, thus highlighting both the relevance of meronyms to object classification, as well as HOLMES ability to capture it. }
The code is available at 
\url{ https://github.com/FrancesC0de/HOLMES}.

\end{abstract}

\section{Introduction}
\label{sec:introduction}

In recent years, the application of Machine Learning (ML) models has impacted the most disparate fields of application. In particular, Deep Learning (DL) models called Convolutional Neural Networks (CNNs) have become the de-facto standard approach to tackle Computer Vision (CV) problems, spanning from autonomous driving to image-based medical diagnosis, from satellite observation to advertisement-driven social media analysis~\cite{cv}.

Unfortunately, DL models are black-boxes, as their fully sub-symbolic internal knowledge representation makes it impossible for human developers and users to understand the rationale behind the model decision process. This widely recognized fundamental flaw has multiple negative implications: $(i)$ difficulty of adoption from domain experts~\cite{trust}, $(ii)$ GDPR non-compliance~\cite{gdpr}, $(iii)$ inability to detect learned spurious correlations~\cite{wrong_reasons}, and $(iv)$ risk of deploying biased models~\cite{hcbias}.

\LM{Due due this plethora of issues, the field of eXplainable Artificial Intelligence (XAI) has flourished in an attempt to make these black-box models more understandable from human developers and users}~\cite{guidotti2018survey}.

In the specific case of CV tasks and CNN models, most XAI approaches are based on saliency and produce heatmaps~\cite{gradcam}, quantifying explanations in the form of {\em this image depicts a cat because of the highlighted region}. On the one hand, this approach can be sufficient to spot wrong correlations when the heatmap focuses on the wrong portion of the image, such as the background. On the other hand, a reasonably-placed heatmap is not a sufficient guarantee that the DL model is in fact implementing the desired task, and we argue that these shallow explanations are not enough for a human user to fully trust the algorithmic decision, nor for a developer to sufficiently debug a model in order to assess its learning progress. These approaches provide context-less label-level heatmaps: ironically, they pair deep models with shallow explanations. Conversely, when asked to justify an image-classification task, humans typically rely on the holonym-meronym (whole-part) relationship and produce part-based explanations, e.g. {\em this image depicts a cat (holonym), because there are pointy ears up there and a tail there, etc. (meronyms)}.

There is evidence that CNNs are capable of learning human-interpretable concepts that, although not explicitly labelled in the training set, are useful to detect classes for which labels are provided; for instance, scenes classification networks learn to detect objects present in scenes, and individual units may even emerge as objects or texture detectors \cite{Bau_2020}. At the same time, CNNs were shown to take shortcuts, relying on contextual or unwanted features for their final classification \cite{Geirhos_2020}; other works found CNNs being over-reliant on texture, rather than shape, for their final classification \cite{geirhos2018imagenet}. In this work, we tackle the important issue of how, and to what extent,  post-hoc explanations can be linked to  underlying, human-interpretable concepts implicitly learned by a network, with minimal effort in terms of annotation and supervision. \\

{\bf Our Research Question is therefore the following:\\
{\em can we decompose the given label (holonym) into a set of related concepts (meronyms), and provide component-level explanations for an image classification 
\LM{DL} model?}}\\

In this paper we propose HOLMES (HOLonym-MEronym based Semantic  inspection), a novel XAI technique that can provide explanations at a low-granularity level.

Given an input image of a given class, its parts (meronyms) are extracted from a Knowledge Base through the holonym-meronym (whole-part) relationship. Images depicting each part are either extracted from a densely annotated dataset or collected through Web scraping, and then used to train a meronym model through transfer learning. The resulting model is therefore a part dectector for the component of the image. The application of XAI techniques on the meronym model can thus
produce part-based explanations.
HOLMES can therefore highlight the occurrence and locations in the image of both labelled objects and their parts.
We evaluated our approach through insertion/deletion/preservation metrics, showing how the parts highlighted by our approach are crucial for the predictions.

The rest of the paper is organized as follows. In Section \ref{sec:rel_work} we connect the proposed technique with other existing approaches in the XAI literature. In Section \ref{sec:meth} we go through the core concepts behind HOLMES. In Sections \ref{sec:exp_set} and \ref{sec:results}, we report experimental validation of the HOLMES pipeline. 

Finally, in Sections \ref{sec:discussion} and \ref{sec:conclusions} we discuss advantages and limitations of the proposed approach, as well as future studies we plan to conduct to enhance HOLMES capabilities.

\section{Related work}
\label{sec:rel_work}

\subsection{Feature Extraction and Transfer Learning}

Deep Convolutional Neural Networks (CNNs) have been the de-facto standard models for computer vision in the last years~\cite{cv}. These models typically encompass a number of convolutional layers, which act as feature extractors, followed by dense layers used for classification. The major drawback of these models is that, due to the large amount of parameters, training from scratch requires a vast amount of data and computational resources~\cite{vgg}. A common technique exploited to circumvent this problem is {\em transfer learning}~\cite{transfer}, in which a model developed for a task is reused as the starting point for a model on a second task. The typical approach for CV tasks is to select  a CNN that was pre-trained on the standard dataset of Imagenet~\cite{imagenet}, and reset and re-train the last dense layers on the new task. 

The underlying intuition of this approach is that CNNs learn a hierarchy of features in convolutional layers at different depths, starting from Gabor filters in the first layers to complex shapes in the last ones~\cite{transfer}.

\subsection{Interpretable and Explainable Machine Learning}

The eXplainable Artificial Intelligence (XAI) research field tackles the problem of making modern ML models more human-understandable. XAI approaches typically belong to one of two paradigms, namely, interpretability and post-hoc explainability\cite{arrieta2020explainable,guidotti2018survey}.
Interpretable ML models are designed and trained in order to be, to some degree, passively transparent - that is, so that comprehensible information about the inner logic of the model is available without the application of other algorithms. Instead, explainability typically is performed {\em a posteriori} - it is a process that takes place after the ML model has been trained, and possibly even deployed. Explainability techniques apply external algorithms to the ML model in order to extract human-understandable information about the decision process that was produced by the training process.

Explainability methods can be further classified according to two orthogonal binary attributes: local/global and model agnostic/aware. 
Local methods provide an explanation for a single data point, while global methods aim to explain the behavior of the model as a whole, e.g., providing a joint explanation for all data points in the dataset. 
Model-agnostic methods can explain indifferently any type of black-box model, regardless of their typology or architecture, accessing input and outputs only. For instance, they could be applied even if the source code of the ML model is obfuscated or can be only accessed through APIs, provided that those can be invoked at will. Conversely, model-aware (also called model-specific) models exploit (and require access to) internal details of the black-box, such as gradients, and are therefore developed for specific kinds of ML models.

\subsection{XAI for Computer Vision }

Arguably the two most famous XAI approaches are LIME~\cite{lime} and SHAP~\cite{shap}, both being local and model-agnostic. An important counterpoint in the field is the concept of global and model-specific approaches, as exemplified by TCAV~\cite{kim2018interpretability}. This methodology allows for global interpretability, focusing on understanding high-level concepts used by the model across a broad set of inputs. However, for the specific task of computer vision, most approaches are model-aware and based on saliency. 

When explaining image classification models, saliency methods compute pixel-level relevance scores for the model final output. These scores can be visualized as heat-maps, overlaid on the classified images, in order to be visually inspected by humans.
One of these approaches is the Gradient-weighted Class Activation Mapping (Grad-CAM)~\cite{gradcam}, a model-aware, local, post-hoc XAI technique.
Grad-CAM uses the gradient information flowing into the
last convolutional layer of the CNN to assign importance
values to each neuron for a particular decision of interest, such as a target concept like {\em dog}.
By visualizing the positive influences on the class of interest (e.g., \textit{dog}) through a global heatmap, Grad-CAM provides insight into which regions of the input image are 'seen' as most important for the final decision of the model. By overlaying this heatmap onto the input image, Grad-CAM facilitates a deeper understanding of the correlation between specific image features and the final decision.

Saliency maps methods such as Grad-CAM ask {\em where} a
network looks when it makes a decision; the network dissection approach takes a step further and asks {\em what} a network is looking for. 
In~\cite{Bau_2020}, the authors find that a trained network contains units that correspond to high-level visual concepts that were not explicitly labeled in the training data. For example, when trained to classify or generate natural scene images, both types of networks learn individual units that match the visual concept of a {\em tree} even though the network was never taught the tree concept during training. The authors investigate this phenomenon by first identifying which individual components strongly correlate with given concepts (taken from a labelled segmentation dataset), and then turn off each component in order to measure its impact on the overall classification task. Following this line of investigation, \cite{zhou2018interpretable} seeks to distill the information present in the whole activation feature vector of a neural network’s penultimate layer. It achieves this by dissecting this vector into interpretable components, each shedding light on different aspects of the final prediction. Our work differs from the network dissection literature in the following ways: (i) we allow for representations of concepts that are scattered across neurons, without forcing them to be represented by a single computational unit; (ii) \LM{we do not require additional, domain-specific ground truth sources, relying instead on web scraping and general purpose-ontologies}
and (iii) we do not focus on the specific scene recognition task, embracing instead the part-of relationships of labels in the more general image classification task.

\subsection{Ontologies and Image Recognition}
Ontologies, and structured representation of knowledge in general, are typically ignored in most DL for image processing papers~\cite{cv}. However, there are notable exceptions where efforts have been made to merge sub-symbolic ML models together with ontologies.

In~\cite{Ghidini}, the authors leverage the fact that ImageNet labels are WordNet nodes in order to introduce quantitative
and ontology-based techniques and metrics to enrich and compare different explanations and XAI algorithms. For instance, the concept of semantic distance between actual and predicted labels for an image classification task allows to differentiate a labrador VS husky misclassification as milder with respect to a labrador VS airplane case.

In~\cite{rodriguez}, the authors introduced a hybrid learning system designed to learn both symbolic and deep representations, together with an explainability metric to assess the level of alignment of machine and human expert explanations. The ultimate objective is to fuse DL representations with expert domain knowledge during the learning process so it serves as a sound basis for explainability. 

Among the global methods for explainability, TREPAN~\cite{trepan} is able to distill decision trees from a trained neural network. By pairing an ontology to the feature space, the authors use the ontological depth of features as a heuristic to guide the selection of splitting nodes in the construction of the decision tree, preferring to split over more general concepts.

\section{Methodology}
\label{sec:meth}
\begin{figure*}[ht]
\centering

\includegraphics[width=0.99\textwidth]{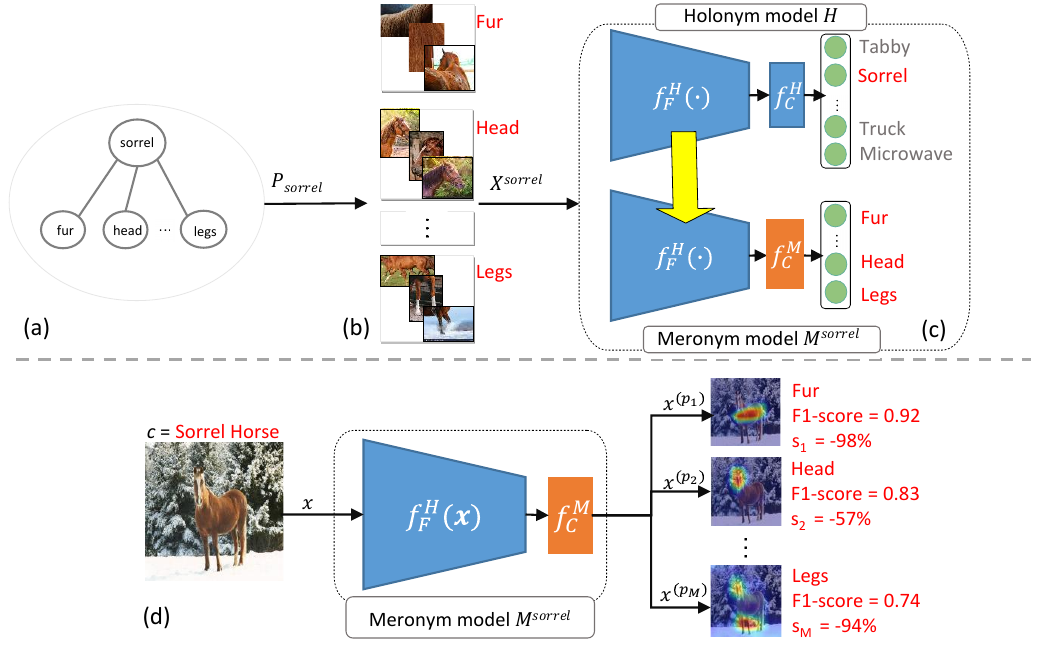}
\caption[HOLMES pipeline]{HOLMES pipeline. Given an input image of class $c$, its  parts (meronyms) are extracted from a Knowledge Base (a). Images depicting each part are either extracted from a densely annotated dataset or collected through Web scraping (b), and then used to train a meronym model by exploiting, through transfer learning, the implicit knowledge embedded in the original holonym model (c). The meronym model then produces  part-based explanations, highlighting the most relevant parts for the class prediction (d). }
\label{fig:pipeline}
\end{figure*}

The proposed method, Holonym-Meronym based Semantic inspection (HOLMES), is a post-hoc approach that aims to explain the classification given by a CNN image classifier to an image in terms of its parts.
It is indeed a model-dependent method, specifically tailored for CNNs.
Hence, HOLMES takes as input the image whose class has been predicted, recovers its meronyms, and provides an explanation in terms of its parts.

\subsection*{Problem Formulation}

\newcommand{\Hfun}{\mathcal{H}}
\newcommand{\Mfun}{\mathcal{M}}
\newcommand{\ImVec}{\mathbf{x}}

Let us define the image classifier as a function $\Hfun: \ImVec \in \mathbb{R}^{h\times w \times ch} \mapsto c \in \mathcal{C}$, where $\ImVec$ is an input image with dimensions $h\times w \times ch$ and $\mathcal{C}$ is the set of image classes.
Lets assume that $\Hfun$ is a CNN that can be expressed as a combination of two functions,
a feature extractor $f^\Hfun_F: \ImVec \mapsto \mathbf{f}$ and a feed forward classifier $f^\Hfun_{C}: \mathbf{f} \mapsto c$, where $\mathbf{f}$ is a feature vector. 
Let us define a holonym-meronym relationship mapping $\mathrm{HolMe}: c \in \mathcal{C} \mapsto \{p_1, ..., p_n\}$ meaning that a whole object of class $c$, i.e. a holonym, is made of its parts, or meronyms $P_c = \{p_1, ..., p_n\}$.
The goal of HOLMES is to explain the classification $\Hfun(\ImVec) = c$ through the meronyms $P_c$, by highlighting which of the meronyms

are important for the classification result.
The explanation takes the form of a function $\xi: \ImVec \mapsto \{(\ImVec^{(p_i)}, s_i),~ p_i \in P_c\}$, where $\ImVec^{(p_i)}$ is a saliency map that highlights the meronym $p_i$ in the original image $\ImVec$, and $s_i$ is an explanation score associated to $\ImVec^{(p_i)}$.

HOLMES solves this problem by training
a new meronyms classifier $\Mfun^c: \mathbf{\ImVec} \mapsto p_i \in P_c$ as a combination of the same feature extractor $f^\Hfun_F$ that is part of the image classifier $\Hfun$ and a new feed forward classifier $f^\Mfun_{P_c}: \mathbf{f} \mapsto p_i \in P_c$. 
The meronyms classifier is then

used to determine which parts $p_i$ are present in the input image $\ImVec$ and to create saliency maps that correspond to these parts. Finally, each saliency map $\ImVec^{(p_i)}$ is used to create a mask on $\ImVec$, which is then classified by $\Hfun$, and the drop of the classifier confidence in the class $c$ is used to determine the importance of the selected parts.

The HOLMES pipeline comprises the following steps:

\begin{enumerate}[A)]
    \item \textbf{Meronyms Extraction}: given an image $\ImVec$ and its predicted class $c$, this first step consists of retrieving the list of the object parts $P_c$ (Fig~\ref{fig:pipeline}(a)). 
    \item \textbf{Meronyms Image Data Collection}: once the object parts list $P_c$ is available, a distinct dataset for each part shall be created in this second step (Fig~\ref{fig:pipeline}(b)).
    \item \textbf{Meronyms model Training}: in this third step, the auxiliary meronyms models $\Mfun^c$ are trained to recognize the object parts $P_c$ by exploiting the knowledge (about the parts) embedded in the original CNN  $\Hfun$ (Fig~\ref{fig:pipeline}(c)).
    \item \textbf{Explanations}: in this last step, a set of part-based explanations is produced, highlighting those parts which are most relevant for the class prediction (Fig~\ref{fig:pipeline}(d)).
\end{enumerate}

\subsection{Meronyms Extraction}

The first step of the pipeline consists of
constructing the holonym-meronym relationship mapping $\mathrm{HolMe}: c \in \mathcal{C} \mapsto P_c$ by
retrieving the visible parts $P_c$ associated to the holonym concept $c$.

Hence, HOLMES relies on external Knowledge Bases (KBs) which include part-of relationships, i.e. containing class concepts (e.g., camel, horse, etc.) and their respective list of parts (e.g., head, legs, etc.), along with information about their visibility. Thus, for obtaining the parts of an holonym concept, HOLMES queries one selected knowledge base for the desired holonym concept and results its associated visible meronyms.

Concepts that are not present in the chosen reference KB

are mapped to the respective WordNet\cite{wordnet} ontology concepts, and the \emph{hypernym/hyponym relationship} is exploited: the WordNet semantical hierarchy is climbed back up to the first hypernym (i.e., a broader class, like {\em bird} for {\em seagull}) which occurs in the reference KB, and its associated (more generic) parts are then assigned to the initial holonym concept.

The meronyms extracted in the previous step are then divided in two categories: \emph{hyper-meronyms} and \emph{hypo-meronyms}.
Given a generic list of meronyms, $P = \{p_1,p_2, …, p_n\}$, hypo-meronyms are parts whose visual space is completely within any other part in $P$. The other parts whose visual space is not completely within any other part in $P$ are the hyper-meronyms.
For instance, for the holonym concept \emph{cat}, the hyper-meronyms would be \emph{head, legs,
feet, tail}, given that none of them is visually contained in any other part, but rather, can only contain hypo-meronyms (e.g., \emph{mouth, whiskers}, etc.).

For the final list of meronyms, only the hyper-meronyms are retained while 
hypo-meronyms are discarded.
The final list of parts is thus defined as $P_c = \{p_1,p_2, …, p_n\}$.

\subsection{Meronyms Image Data Collection}

Once the part set $P_c$ for the target class $c$ is available, the next step is to create a dataset

$X^c = \{ (\ImVec_0, y_0), ..., (\ImVec_n, y_n)  \}$ 
where $\ImVec_0, ..., \ImVec_n$ are images corresponding to parts $y_0, ..., y_n \in P_c$.

HOLMES can rely on a pre-existing labelled dataset, or it can exploit web image scraping to incrementally build a dataset for each meronym. In this scenario, HOLMES queries different web search engines for each part, by prefixing the holoynm first (e.g. {\em sorrel fur}, {\em sorrel head}, etc.) and downloads the associated images from each of those engines. 

Due to the limited reliability of search  engine results (discussed in Section~\ref{sec:exp_set}),

some obtained images could be still extraneous to the desired part concept. Moreover, duplicates could be present in the scraped parts’ datasets. For these reasons, HOLMES integrates two additional sub-steps:
\begin{enumerate}
    \item \textbf{deduplication}: duplicates are detected by means of the pHash\cite{phash} hash-based deduplication method, then they are removed from each meronym dataset.
    \item \textbf{outlier removal}: meronyms images are mapped to a feature vector representation (e.g. using the output of the feature extractor or the activations of one of the feedforward layers of the classifier).

    The feature vectors are then fed to an outlier detection algorithm. The detected outliers are then removed from the meronym dataset.
\end{enumerate}

\subsection{Meronyms model Training}

The training phase is the core of the HOLMES method. In this step the concept parts are visually learned, so that they can later be provided as explanations. This is achieved by training an auxiliary CNN model $\Mfun^c$, trained and evaluated on the collected meronym dataset $X^c$ (training and evaluation are performed in disjoint sets).

Let us recall that the goal of HOLMES is to explain the target holonym CNN $\Hfun(\ImVec) = \hat{y}$, where $\ImVec$ is an holonym image of class $c$ and $\hat{y}$ its predicted class. Let us also recall that the CNN can be expressed as a combination of two functions $\Hfun(\ImVec) = f^\Hfun_{C}(f^\Hfun_F(\ImVec))$, where $f^\Hfun_F( \cdot )$ is a feature extractor, and $f^\Hfun_{C}( \cdot )$ is a feedforward classifier. 
Previous works already demonstrated
that the units contained in the last convolutional layers of a CNN tend to embed objects, and more specifically, objects parts as well \cite{gonzalezgarcia2017semantic,Bau_2020}, and HOLMES leverages on this fact to learn the parts by defining $\Mfun^c(\ImVec) = f^\Mfun_{P_c}(f^\Hfun_F(\ImVec))$, where the feature extraction $f^\Hfun_F( \cdot )$ is shared among the holonym $\Hfun$ and meronym  $\Mfun^c$ models, whereas a feedforward classifier $f^\Mfun_{P_c}( \cdot )$ is trained anew for each class $c$ and each part list $P_c$.  

The idea is to learn the parts concepts by using the same features learned by the original reference CNN model $\Hfun$, such that the base knowledge for learning both the concept parts and the concepts themselves would be the same: effectively, HOLMES relies on transfer learning\cite{yosinski2014transferable} for learning objects parts. Under the reasonable assumption that characteristic object parts, and consequently their associated features, are useful for the classification of the whole object itself, the same units which activate in the presence of the parts will also activate in presence of the object. For instance, a unit activating in the presence of a wheel, will also be likely to activate in the presence of a wheeled vehicle like a car. Hence, training $\Mfun$ by keeping the feature extractor   $f^\Hfun_F$ intact will later allow us to understand if the knowledge about the parts was already available and embedded in the original model $\Hfun$.
Specifically, the feature  maps obtained in the presence of the individual parts will be useful to create a visual explanation for the (holonym) predictions of the original model.

A held-out test set is used to calculate the per-part calibrated F1-score~\cite{calibration} to determine to which degree each part was learned and distinguished among the others. 
The F1-score is calibrated to be made invariant to the class prior,
so that performances of models trained on different numbers of meronyms can be compared.

\subsection{Explanations}

At the end of the previous step a trained meronyms CNN model $\Mfun^c$ is obtained. For any input holonym image $\ImVec$, this model outputs a set of prediction scores $Y_p = \{y_{p_1},…,y_{p_n}\}$, where $n$ is the the number of parts the model was trained on, and $y_{p_1},…,y_{p_n}$ are the scores produced for each different part. Hence, by feeding the network with an holonym sample (such as a car image), a score about each of its parts (e.g., wheel, bumper, etc.) will be produced. Intuitively, the output scores reflect {\em how much} of each part the network sees in the input holonym image. Exploiting the fact that the network can ‘see’ the part concepts within an holonym image sample, we can look {\em where} the network exactly sees the parts, i.e., in which portion of the input image.

Specifically, the visualization of each part in the holonym image is obtained trough the state-of-the-art saliency method Grad-CAM\cite{gradcam}. After obtaining a saliency map $\ImVec^{(p_i)}$ related to each part that the network can recognize, each saliency map $x^{(p_i)}$ is thresholded into a binary segmentation mask $m^{(p_i)} \equiv ( x^{(p_i)} \geq T^{(p_i)} ) $, where $T^{(p_i)}$ is set to the $q^{th}$ percentile of the corresponding saliency map pixel distribution. We later feed the same input holonym image into the original CNN model, and verify whether each part is fundamental for the original network prediction, by ablating one part from the image at a time based on the meronyms masks $m^{(p_i)}$.  By observing the score drop for the original predicted holonym class label (calculated in percentage, with respect to the original holonym score), we can determine how much the removed meronym was important in order to predict that class label: the more consistent the drop, the more significant the visual presence in the image of the part would be for the original model.

At this point, the input image $\ImVec$ is associated to a set of saliency maps $\ImVec^{(p_i)}$ for each part $p_i \in P_c$, and each saliency map is associated to a score drop $s_i \in S=\{s_1, s_2, ..., s_n\}$.

Additionally, the per-part calibrated F1-score previously computed is used to measure the reliability of the part identification.
We assume that a meronyms model which had difficulties to learn and distinguish a part, would have consequently achieved a low F1-score for that part.
Hence, the parts whose holonym score drop $s_i$ exceeds a threshold $T_s$ and whose meronyms model are above a F1-score threshold $T_{F1}$ are provided as part-based explanations for the original model prediction, as it would mean that those parts are both correctly detected by the meronym model and deemed relevant for the classification of the holonym.

\section{Experimental Settings}
\label{sec:exp_set}

\begin{figure*}[th]
\centering

\subfigure[PASCAL-Part]{\includegraphics[width=.99\textwidth]{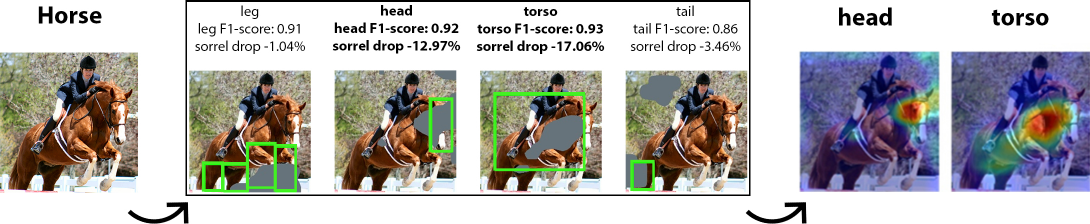}} \\
\subfigure[ImageNet]{\includegraphics[width=.99\textwidth]{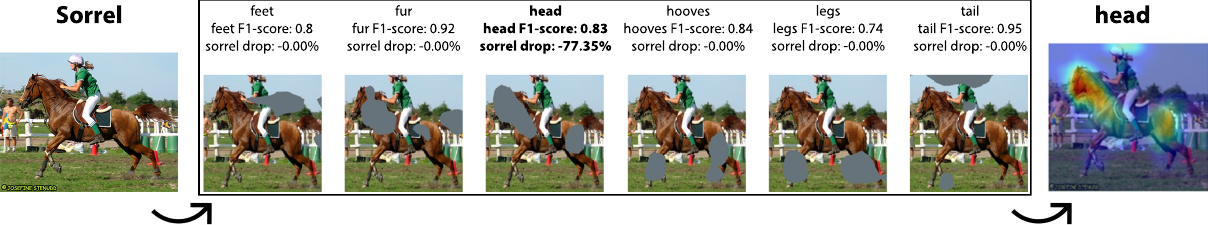}} \\
\caption[HOLMES Explanation]{HOLMES Explanation example for the {\em horse} class -- PASCAL-Part (a) and {\em sorrel} class -- ImageNet (b). For each part, the corresponding ablation mask (grey), the per-part calibrated F1-score and the holonym score drop are shown. For PASCAL-Part, the ablation masks are compared against the ground truth bounding boxes (green). The final heatmap(s) show the part-based explanations. Two and one part are included in the explanations for examples (a) and (b), respectively, as they exceed both the holonym score drop threshold $T_s$ (0.1) and the calibrated F1-score threshold $T_{F1}$ (0.7).}
\label{fig:explanation}
\end{figure*}

HOLMES can generate part-based explanations for any model that can be expressed as a feature extractor and a feed forward classifier.
In our experiments, we explain the outputs of a VGG16~\cite{vgg} image classifier pre-trained on the ImageNet~\cite{imagenet} dataset. In this section, we describe the application of the HOLMES pipeline in two different experimental settings to explain the outputs of the VGG16 model.
In the first experiment, we exploit bounding boxes for objects and their parts from the PASCAL-Part dataset to explain and validate the results.
In the second one, we first build a part-based mapping for many ImageNet classes, then we use scraping to build a dataset for training the meronyms models, and finally we generate part-based explanations and evaluate the results with insertion, deletion and preservation curves. Examples of HOLMES explanation on both datasets are provided in Fig.~\ref{fig:explanation}.

\subsection{PASCAL-Part dataset}
\label{ssec:setting-pascal}

The PASCAL-Part~\cite{pascalparts} dataset contains additional annotations over the PASCAL VOC 2010 dataset, i.e., bounding boxes for objects and their parts, that can be used as the holonym-meronym relationship mapping ($\mathrm{HolMe}$). We held out a set of 50 images for each of the selected holonyms and their corresponding cropped meronym images that are used exclusively to evaluate the HOLMES explainability and part localization performance. The other images were used in the meronym models $\mathcal{M}^c$ training.

\textbf{Meronyms extraction settings:}
The twenty classes in the Pascal VOC 2010 dataset can be divided in four macro-classes: (1) person, (2) animals (bird, cat, cow, dog, horse, sheep), (3) vehicles (aeroplane, bicycle, boat, bus, car, motorbike, train) and (4) indoor objects (bottle, chair, dining table, potted plant, sofa, tvmonitor).
Two of the classes (person and potted plant) have no corresponding class in the ImageNet 1000 classes and thus are discarded.
Other five classes were discarded because they do not have part-based annotations (boat, chair, dining table, sofa, and tvmonitor).
For each of the 13 remaining classes, the respective meronyms $P_c$ were extracted from the PASCAL-Part parts list.
For the six animal classes we performed 
hyper-meronym selection as in \cite{interpretableCNN}, and
for the remaining classes we selected the hyper-meronyms by majority voting. The final $\mathrm{HolMe}$ mapping is thus:
\begin{itemize}
\small
\item $ P_{bird} = P_{cat} = P_{dog} = P_{horse} = $ \{head, torso, leg, tail\}
\item $ P_{cow} = $ \{head, torso, leg, horn\}
\item $ P_{sheep} = $ \{head, torso, leg\}
\item $ P_{aeroplane} = $ \{stern, wheel, artifact wing, body, engine\}
\item $ P_{bicycle} = $ \{saddle, wheel, handlebar\}
\item $ P_{motorbike} = $ \{saddle, wheel, handlebar, headlight\}
\item $ P_{car} = P_{bus} = $ \{window, wheel, headlight, mirror, door, bodywork, license plate\}
\item $ P_{train} = $ \{coach, locomotive, headlight\}
\item $ P_{bottle} = $ \{body, cap\}
\end{itemize}

\textbf{Meronyms image extraction settings:}
Once the holonym-meronym relationship mapping is defined, we extract the images to train the meronyms models. For each holonym image, we retrieved the bounding box coordinates associated with their parts $P_c$, and cropped the holonym image accordingly to produce a set of meronym images.

\LM{
In order to obtain images compatible with the square VGG16 input, while preserving the aspect ratio and shape of each part, before cropping we extended the bounding box in the $x$ or $y$ direction to obtain a square crop, with the constraint of not overlapping other bounding boxes in the same image. Then, if a 1:1 aspect ratio was not completely reached, we applied padding to get a final square crop.}

Moreover, the number of crops extracted for each meronym might be very different, e.g., for a {\em horse} meronym, there are more {\em leg} parts with respect to just one {\em head}. To avoid high class unbalance, we used data augmentation to balance the number of samples in each meronyms class. Specifically, we applied both random rotation and random shear, and one among the gaussian blur, emboss, and gaussian noise transformations to each cropped image.
Finally, for each holonym class, the extracted meronyms samples were split into training/validation/test folds with ratios of 0.81/0.09/0.1.

\textbf{Training and Explanations settings:}
For each holonym, we built a separate meronym model $\Mfun^c$ and we retrained a feed forward classifier $f^\Mfun_{P_c}( \cdot )$ with the same structure of the original VGG16 classifier using Cross Entropy Loss.
Common data augmentation techniques were employed: horizontal flipping, rotation, cropping, color jittering, and random gray scale. Each meronym model was trained for 100 epochs, with a batch size of 64 and learning rate of 0.001 (determined experimentally). 
Early stopping policy with patience set to 5 was used to avoid overfitting.

Regarding the explanations, the activations of the last convolutional layer of VGG16 were used to produce the Grad-CAM meronyms heatmaps, which are then binarized 
using a threshold $T^{(p_i)}$.
We found $T^{(p_i)} = 83^{th}$ percentile 
by performing a grid search upon the $[75, 90]$ percentile values and by finding the best trade-off between different causal metrics performance (described in the Evaluation section) on the whole PASCAL-Part training set, comprising all training holonym image samples. 
The masked pixels were ablated by replacing with the gray RGB value (as the ImageNet mean pixel is gray) for retaining the natural image statistics~\cite{samek2017evaluating}. Finally, the $T_s$ and $T_{F1}$ thresholds were set to 10 and 0.7, respectively.

\textbf{Evaluation settings:}
The whole HOLMES pipeline was run and tested upon each validation image sample associated to each selected class.
The meronyms localization performance was measured by computing the per-pixel AUC score of the HOLMES meronym heatmap versus the same meronym ground truth. To calculate this metric, each pixel was assigned the corresponding heatmap value as score: true positive pixels were those belonging to the actual part (i.e., falling within the bounding box), while the remaining pixels were labeled as false positives. The performance is compared to the per-pixel AUC score of the Grad-CAM holonym heatmap as baseline. 
Moreover, the faithfulness of HOLMES explanations was assessed by means of common causal metrics based on the deletion/insertion/preservation curves ~\cite{rise,preservation}. 

\LM{As mentioned before, HOLMES produces a set of part-based explanations, which are obtained by computing a set of saliency maps $\mathbf{X}^{(P_c)} =  \{\ImVec^{(p_i)},~ p_i \in P_c\}$, each associated a specific part $p_i$.  
However, to assess the global quality of such explanations, all part-based saliency maps need to be merged into a unique heatmap,
comprising all parts. Given the set of saliency maps $ \mathbf{X}^{(P_c)} $, and the corresponding score drops $ S=\{s_1, s_2, ..., s_n\} $ associated with the ablation of each part, the HOLMES global heatmap is obtained through a weighted linear combination of the part-based saliency maps. First, normalized score drops $ Z=\{z_1,..., z_n\}$ are calculated by dividing each score drop by the L1-Norm of $S$. Then, the global heatmap is obtained by summing each weighted heatmap element-wise: $G=\sum_{i \in {n} \ImVec^{(p_i)}z_i}$. This weighting scheme emphasizes parts whose ablation causes a significant holonym class score drop.}

After having obtained the global heatmap $G$ for an input image in this way, it is so possible to use causal metrics such as the areas under the insertion~\cite{rise}, the deletion~\cite{rise} and the preservation curves~\cite{preservation} to assess the
overall quality of the part-based explanations, whose information is combined into $G$. These metrics were computed for all held out PASCAL-Part validation images. It is crucial to note that, distinct from simply replicating Grad-CAM results, our global heatmap stresses the pivotal role of part-based explanations, serving as an integral instrument to appraise the global effectiveness of the part-based explanations.

\subsection{ImageNet}
HOLMES can be also applied in scenarios where  part-level annotated datasets are not available. In this alternative case, we leverage ontologies and image scraping to build the necessary meronym datasets. In particular, we exploit the connection between Imagenet\cite{imagenet} labels and WordNet\cite{wordnet} nodes in order to retrieve a list of parts of the object-label, relying on the holonym-meronym (whole-part) relationship. 

\textbf{Meronyms extraction settings: }
Across the ImageNet 1000 class concepts, 81 of them were selected and treated as holonym classes. The selected holonym classes belong to two main categories:
\begin{enumerate}
    \item Medium- or large-size animals
    \item Medium- or large-size man-made objects 
\end{enumerate}

The size constraint is necessary to obtain acceptable training sets. In fact, the smaller the holonym (e.g., bugs in the animals category), the more troublesome it becomes to retrieve  images of distinct parts by querying web search engines. Specifically, when querying for such parts, the engines tend to return images of the whole holonym concept instead (e.g., the whole butterfly when querying for a butterfly head). This would consequently result in meronyms datasets very similar among each other and with a strong visual overlap, thus greatly hindering the associated meronym model performance.  

Therefore, for each of the 81 classes, the respective meronyms were extracted from the Visual Attributes for Concepts (VISA)\cite{visa} dataset. Hyper-meronyms were further extracted by manual filtering: the meronyms obtained from the ontology were manually categorized into hyper-meronyms and their respective hypo-meronyms. In this way, for each occurrence of a hyper-meronym, the associated hypo-meronyms were automatically filtered out. The final $\mathrm{HolMe}$ mapping is available in the Supplementary Material, Section I.A. 

\textbf{Meronyms image scraping settings:} The Google and Bing web search engines were selected and queried for downloading the images.
The number of downloads per part over all the engines was forcibly limited, since the pertinence of the images with respect to the desired part concept naturally decreases as more images are downloaded (e.g., after too many downloads for {\em sorrel head}, an engine would for instance start returning images depicting plants and flowers); a good rule-of-thumb is to limit the download to the first 100 items \cite{ricci2019image,massouh2017learning}. 
Finally, in order to further increase the dataset size for each part, the {\em Visually similar images} function of Google was exploited: for each downloaded image, the most visually similar ones are searched in this way and then added to the parts’ samples.
The download limit, i.e., the number of images to be downloaded for each part by each engine, was set to 40 for Google and 60 for Bing, since Bing showed to be slightly more reliable. 
The visually similar images download limit was instead set to 5.  Duplicates and near-duplicates \cite{morra2019benchmarking} are detected by means of the pHash\cite{phash} hash-based deduplication method. For outlier removal, meronyms images are mapped to a feature vector using the activations of the penultimate FC layer of VGG16\cite{vgg}, which are then given as input to the PCA outlier detection algorithm\cite{pcashyu}, with the outlier contamination rate hyper-parameter set to 0.15.  The scraped data was split into training/validation/test folds with proportions of 0.81, 0.09 and 0.1 respectively.

\textbf{Training and Explanations settings:} The training and explanation steps are carried out with the same settings as detailed for the PASCAL-Part dataset (Section \ref{ssec:setting-pascal}).

\textbf{Evaluation settings:} The global heatmap is evaluated using the insertion, deletion, and preservation curves as detailed for the PASCAL-Part dataset (Section \ref{ssec:setting-pascal}).

\section{Results}
\label{sec:results}

The HOLMES pipeline was quantitatively and qualitatively evaluated in all its steps. Experimental validation aimed at determining i) to what extent HOLMES is able to correctly identify and locate meronyms?, ii) to what extent the classification score can be attributed to individual meronyms and iii) how good are the explanations generated by HOLMES? 

\textbf{RQ1: How well can HOLMES classify and locate meronyms?}

As introduced in Section \ref{ssec:setting-pascal}, the PASCAL-Part ontology contains 13 classes with an average of $\approx 4$ visible parts per class. Following the procedure described in the experimental setting, on average $\sim 750$ sample per meronym were collected ($\sim 1400$ after data augmentation), for a total of 74,772 training samples. For the ImageNet dataset, 81 classes were selected, with an average of $\approx 7$ visible parts per image. Thus, web scraping was performed for a total of 559 meronyms, yielding on average $\sim450$, of which 18\% were detected as duplicates and 11\% as outliers, and hence, eliminated. The final average number of images per part is  $\sim320$.

\begin{figure}[tb]
\centering
\includegraphics[width=.75\linewidth]{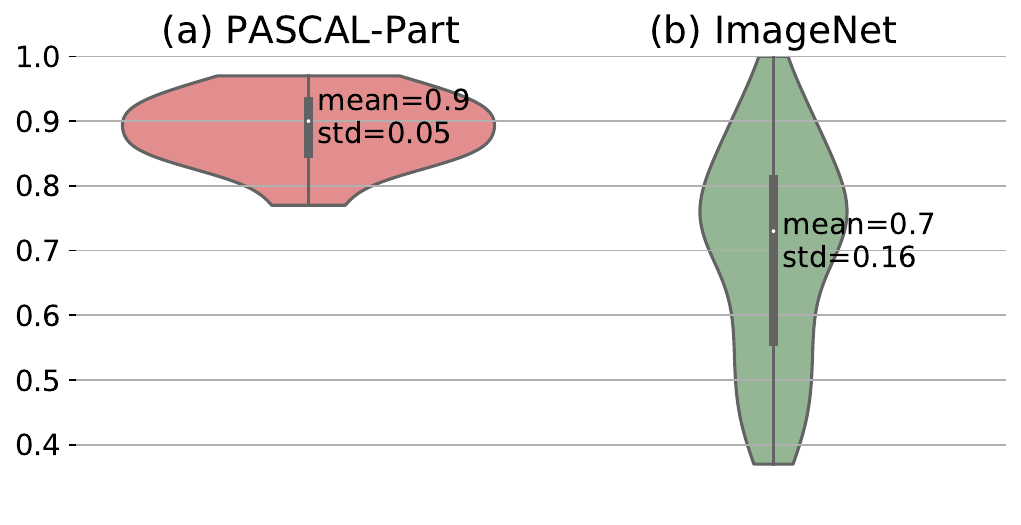}
\caption[F1-score results]{Distribution (violin plot) of the  average per-part calibrated F1-score. 
}
\label{fig:F1_violin}
\end{figure}

First, we assess HOLMES ability to \emph{classify different meronyms} by reporting the distribution of the calibrated F1-scores of the $\Mfun^c$ models, trained upon each training set $X_c$ for each of the selected classes, is reported in  Fig. \ref{fig:F1_violin} for both PASCAL-Part and ImageNet dataset. The average F1-score was good in both cases, but higher on PASCAL-Part (0.9 $\pm$ 0.05 vs. 0.7$\pm$ 0.16). This difference can be attributed, at least partially, to the higher precision of the PASCAL-Part reference standard, for which bounding boxes are available.

\begin{figure}[tb]
\centering
\includegraphics[width=\linewidth]{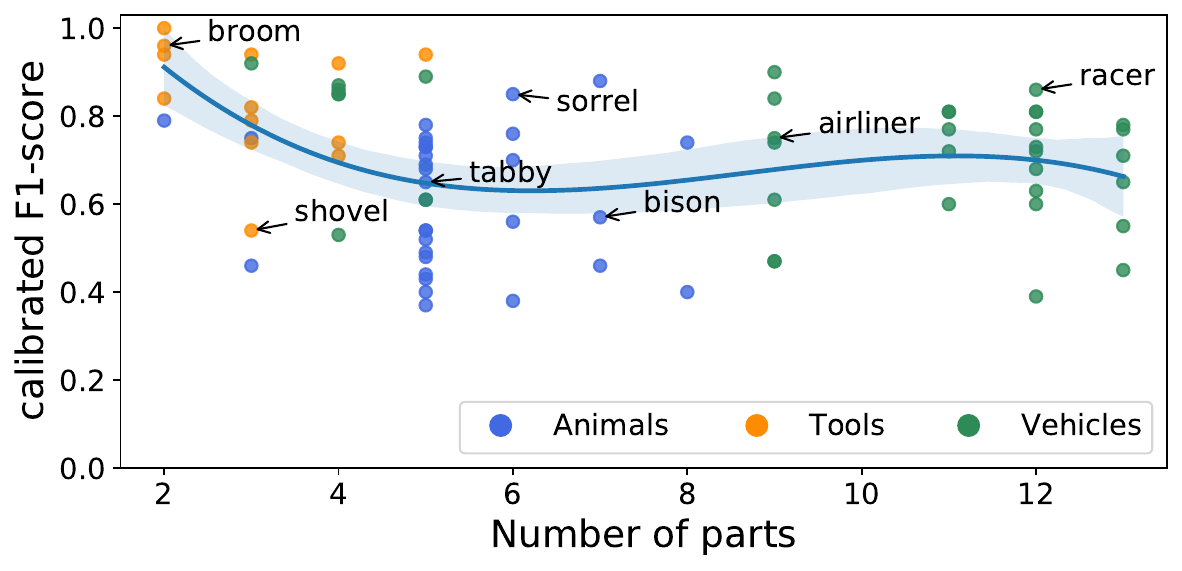}
\caption[F1-score results]{Average per-part calibrated score as a function of the number of parts per holonym class (colored dots represent a holonym, blue line is the mean average per-part F1-score).  
}
\label{fig:F1_dist}
\end{figure}

\begin{table*}[tbh]
\caption {The HOLMES meronyms localization performance is measured by computing the average per-pixel AUC score of each HOLMES meronym heatmap (top row) against the holonym heatmap extracted by Grad-CAM (bottom row).} \label{tab:pixelAUC results}
    \centering
    \resizebox{\textwidth}{!}{
    \begin{tabular}{||ccccccccccccccc||}
\hline
Method & horse & cat  & bird & cow  & dog  & sheep & aeroplane & bicycle & bottle & bus  & car  & motorbike & train & Avg  \\ [0.5ex] 
\hline
HOLMES & 0.77  & 0.74 & 0.8  & 0.77 & 0.74 & 0.75  & 0.74      & 0.76    & 0.67   & 0.75 & 0.68 & 0.74      & 0.71  & \textbf{0.74} \\
Grad-CAM & 0.68  & 0.68 & 0.71 & 0.66 & 0.71 & 0.62  & 0.76      & 0.63    & 0.6    & 0.65 & 0.62 & 0.66      & 0.62  & 0.66 \\
\hline
\end{tabular}
}
\end{table*}

On the other hand, it can be observed from Fig. \ref{fig:F1_dist} how the performance degrades proportionally to the number of parts per class. This is especially evident for the ImageNet dataset, since the ontology is richer with more classes and more parts per classes. Indeed, as the number of parts increases, so does the likelihood that images scraped as belonging to different parts will have some degree of visual overlap, which would negatively impact the performance of the trained $\Mfun^c$ model. At the same time, different class categories tend to be associated with a lower/higher number of  meronyms: for instance, tools tend to have between one and four parts, animals between three and eight, and vehicles more than eight. Thus, we cannot exclude that the category may also play a role either by influencing the quality of the scraping, or the differentiation of the meronyms themselves. 

The HOLMES meronyms localization performance was measured by computing the per-pixel AUC score of each HOLMES meronym heatmap against their PASCAL-Part ground truth bounding boxes.
As a baseline, we could assume that either the meronym could be randomly assigned to any region of the image (in this case, AUC=0.5), or we could focus on the actual holonym as extracted by the Grad-CAM algorithm and assume that the meronym is inside the region of the Grad-CAM holonym heatmap.
This second choice offers a baseline that is harder to beat, but as reported in Table \ref{tab:pixelAUC results},
the HOLMES meronyms explanations consistently localize the parts better and more precisely, compared to the whole Grad-CAM heatmaps which instead localize the entire object.

\textbf{RQ2: To what extent the classification score can be attributed to individual meronyms?}

Having established the ability to classify meronyms, the next step is to evaluate their impact on the holonym classifier $\Hfun$, as exemplified in Fig. \ref{fig:explanation}.  

\begin{figure}[bt]
\centering
\includegraphics[width=.75\linewidth]{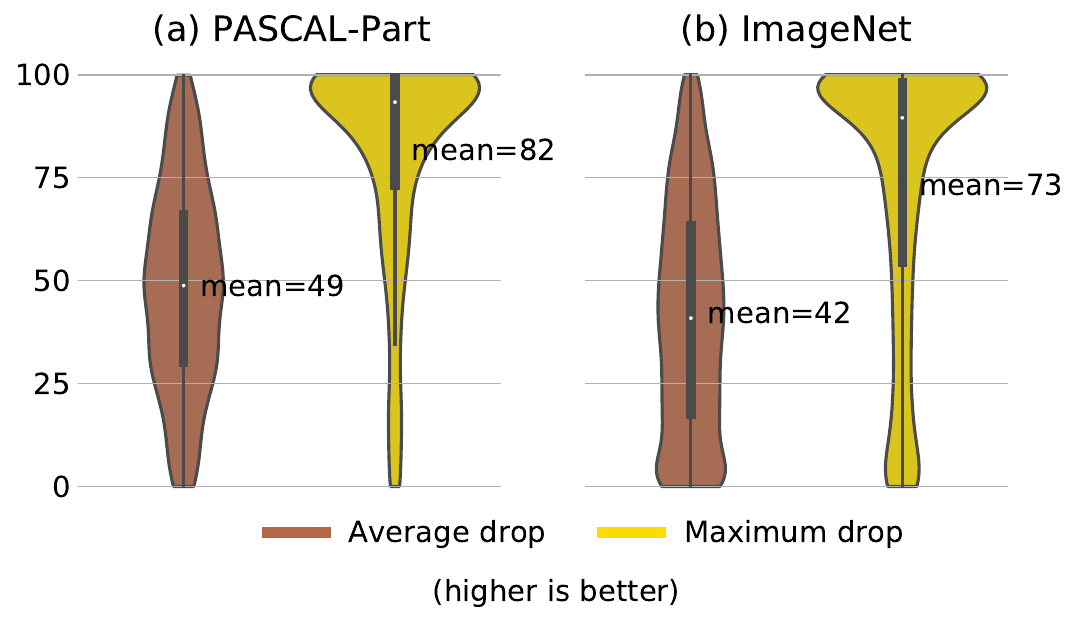}
\caption[Mean drops]{Distribution (violin plots) of the average score drop and maximum score drop (in percentages) per image on the PASCAL-Part (a) and ImageNet (b) validation sets. The score drop is calculated for each image and meronym by ablating the corresponding mask; then, the average and maximum score drop are computed over all meronyms appearing in an image.}
\label{fig:mean_drops}
\end{figure}

\begin{figure*}[!tbh]
\centering
\includegraphics[width=\textwidth]{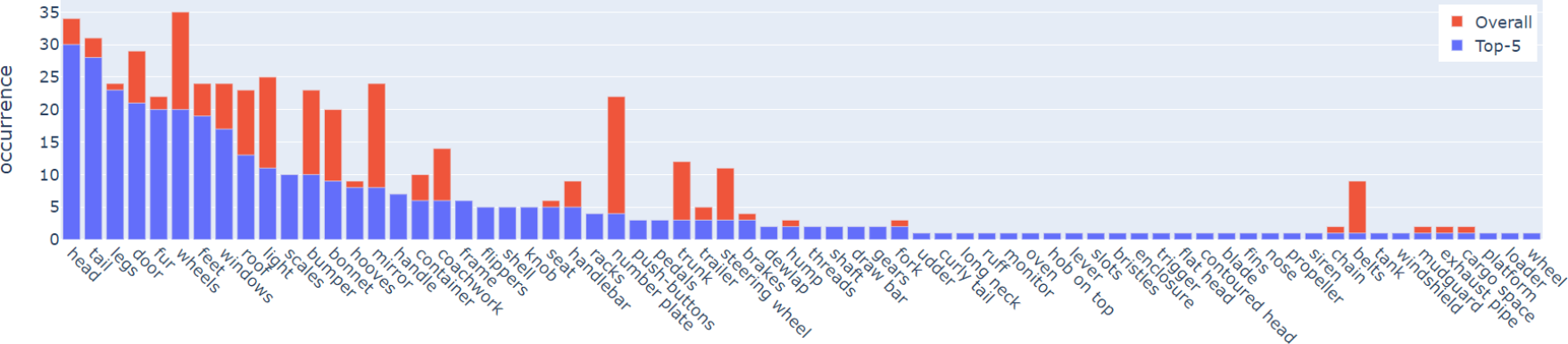}
\caption[Meronyms frequency]{Top-5 meronyms distribution. For each class, the top-5 meronyms are computed (i.e., the meronyms inducing on average the highest score drops).  For each meronym, the total number of associated holonyms (blue) is compared with the frequency they appear in the Top-5 meronym list (red). Meronyms are listed in decreasing order of frequency. }
\label{fig:meronyms_fr}
\end{figure*}

The distribution of the \textit{per-meronym score drop}, i.e., the score drop observed for the holonym when the corresponding meronym is ablated, is reported in Fig.~\ref{fig:mean_drops}. The average score drop is 49\% for PASCAL-Part and 42\% for ImageNet, respectively, meaning that on average, the ablation of a single meronym roughly halves the holonym model confidence.  When considering only the most significant part (i.e., the one associated with the highest score drop for each test image), the score drop increases to 82\% on PASCAL-Part. Hence, in this dataset, individual meronyms have a substantial impact on the classifier output and, in most cases, the classification can almost be fully explained or attributed to a single meronym. On the other side, on ImageNet the mean maximum drop is lower (73\%). In fact, the mean ($\pm$ standard deviation) number of meronyms included in each explanation was $2.28  \pm 2.46$. Examples of explanations with one, two, three or more meronyms are provided in the Supplementary Material. 

\LM{Conversely, when grouping images according to their holonym labels, the mean average drop and mean maximum drop were 50\% and 80\% for PASCAL-Part, and 46\% and 73\% for ImageNet, respectively. While the score distribution in Fig.~\ref{fig:mean_drops} characterizes HOLMES behavior at the instance (image) level, the mean values give some insight into the general properties of the holonym classes. For instance, the mean maximum score drop on PASCAL-Part shows a bimodal distribution, with animal classes having a higher mean maximum drop than vehicles and man-made objects (78.5\% vs. 62.1\%). This entails that, on average, explanations for image belonging to animal classes are highly focused, with only one meronym almost explaining the entire holonym concept, whereas for other classes explanations require the combination of two or more meronyms. On ImageNet, we did not find a large difference between animals (70\%), tools (73\%), and vehicles (74\%); on the other hand, we observed a wider variation between classes, with mean maximum drop ranging from 23\% (zebra) to 95\% (persian cat).  }

Different holonym classes may have an overlapping set of meronyms: this is especially evident for the richer ImageNet ontology. For instance, most of the animal classes have a head and a tail (although each class will be associated with its own training dataset and its own classifier $\mathbf{M}^c$).  Moving from this observation, we sought to understand whether certain parts induced a consistent and substantial holonym score drop (Fig.~\ref{fig:meronyms_fr}). For instance, for classes belonging to the animal category, the meronyms head, tail, and legs frequently cause a consistent drop when ablated from the image. Similarly, for vehicles, the door, wheels and window meronyms are the ones with the highest impact on the holonym class prediction.

\begin{figure*}[tbh]
\centering
\includegraphics[width=\textwidth]{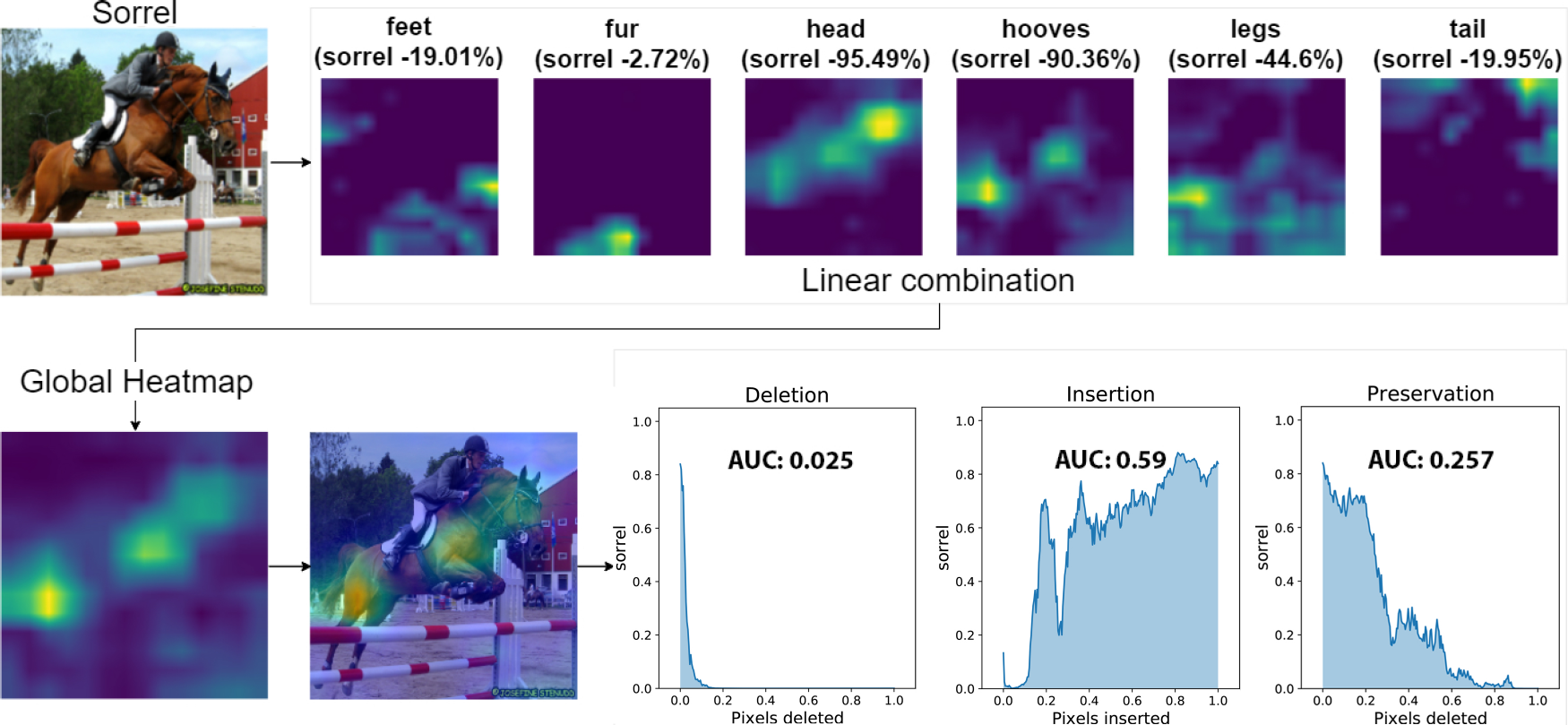}
\caption[HOLMES Global Explanation]{HOLMES Global Explanation. Starting from an input image, the per-part heatmaps and the respective holonym score drops are obtained. Then, by a linear combination of the heatmaps, the global heatmap is obtained, and its quality is measured by means of the insertion/deletion/preservation metrics. 
}
\label{fig:global_exp}
\end{figure*}

\textbf{RQ3: How good are the explanations generated by HOLMES?}

Finally, the overall quality of the generated explanations was evaluated. For the purpose of evaluating against Grad-CAM, which is not designed to provide part-based explanations, part-level heatmaps are linearly combined in a global heatmap, as detailed in Fig.~\ref{fig:global_exp}. The AUC values under the insertion, deletion, and preservation curves are summarized in Table \ref{tab:curves results}.

\begin{table}[tb]
\caption {Deletion/Insertion/Preservation AUCs for HOLMES and Grad-CAM.} \label{tab:curves results}
    \centering
    \begin{tabular}{|p{2.0cm}@{\hspace{5pt}}p{2.0cm}ccc|}
\hline
Dataset      & Method   & Deletion $\downarrow$     & Insertion $\uparrow$      & Preservation  $\uparrow$ \\ 
\hline

PASCAL-& HOLMES   & $0.050\pm0.053$  & $0.487\pm0.269$  & $0.392\pm0.255$ \\
Part             & GradCAM & $0.052\pm0.060$  & $0.505\pm0.277$  & $0.381\pm0.264$ \\
\hline
ImageNet     & HOLMES   & $0.112\pm0.113$  & $0.660\pm0.252$  & $0.538\pm0.257$ \\
             & GradCAM & $0.111\pm0.107$  & $0.684\pm0.242$  & $0.539\pm0.261$ \\

\hline
\end{tabular}
\end{table}

\begin{figure}[!tb]
\centering
\includegraphics[width=.75\linewidth]{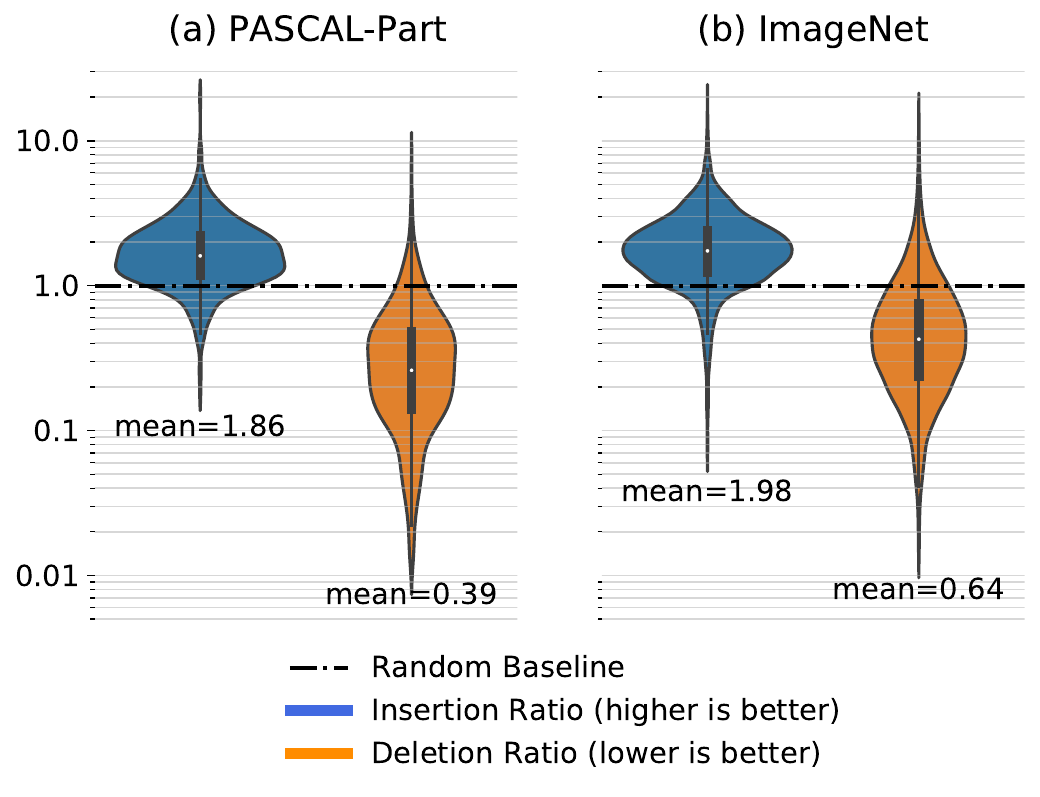}
\caption[Insertion/Deletion Evaluation Summary]{Insertion/Deletion Ratio distribution (violin plot) for the PASCAL-Part (a) and ImageNet (b) datasets. The average insertion ratio (left) and the average deletion ratio (right) are calculated with respect to the random baseline (dotted black line).
}
\label{fig:curves_ratio}
\end{figure}

\LM{
On PASCAL-Part, the HOLMES insertion AUC is, on average, 0.96 times the GradCAM insertion AUC, while the average deletion AUC is 0.95 and the average preservation AUC is 1.03 times the respective GradCAM score. Analogously, on ImageNet, the average insertion, deletion, and preservation AUCs are 0.96, 1.01, and 0.99 times the corresponding GradCAM scores.} 

Additionally, we compared HOLMES  against a random baseline obtained by dividing the images in super-pixels, which are then erased in random order. The random baseline is designed to account for the object scale: in fact, a good heatmap for a small object will yield a lower deletion AUC than an equally good heatmap for a larger object.  As shown in Fig.~\ref{fig:curves_ratio}, \LM{HOLMES metrics are substantially higher than the random baseline, with average insertion AUC 0.58 lower and average insertion AUC 1.77 higher than the baseline.  }

\section{Discussion}
\label{sec:discussion}

Unlike previous methods~\cite{Bau_2020}, HOLMES does not require a densely annotated dataset with pixel-level annotations. Instead, it can be trained using weak annotations either in the form of bounding boxes, such as those available in the PASCAL-Part dataset~\cite{pascalparts}, or relying on the potentiality of web scraping, which drastically reduces the annotation effort, whilst forgoing the limiting closed-world assumption intrinsic to traditional labelled datasets. The effectiveness of web scraping for object recognition has been established in previous works~\cite{massouh2017learning,yao2020exploiting},  which HOLMES capitalizes on and extends, using deduplication and outlier removal 
to reduce noise and increase variety in the training dataset. At the same time, retrieving high quality images for  meronyms, as opposed to holonyms, introduces additional challenges which may impact the quality of the dataset and, thus, the meronym models. Qualitatively, we observed that the pertinence of the retrieved images is generally good, but may decrease depending on the popularity of the meronym as a search term. Another specific challenge is the visual overlap, as it is difficult to find images that precisely isolate one and only one meronym. 

Overall, quantitative evaluation on PASCAL-Part allowed us to conclude that the meronym models are capable of detecting object parts and locating their position within the image. This is achieved by exploiting, without any retraining or fine-tuning, the features learned by the holonym model, thus further supporting the conclusion that knowledge about object parts is implicitly embedded in deep neural networks~\cite{zhou2015object,bau2017network,Bau_2020}.

Partly due to the imperfect background, the ablated mask does not always provide a perfect segmentation of the part itself, as shown in Fig.~\ref{fig:explanation}. In some cases, the ablated masks for different parts could be very similar, especially for meronyms that are physically next to each other. An example is provided by the meronyms {\em legs} and {\em hooves} for the holonym {\em sorrel}, as shown again in Fig.~\ref{fig:explanation}. Less frequently, it may occur that the ablated part may include a portion of the background; for instance, in the case of legs, some terrain or grass may be included. \LM{This may have an impact on the score drop observed when the corresponding part is deleted, and the resulting metrics. }

The average F1-score for most holonyms ranges roughly between 0.6 and 1.0 for ImageNet, and between 0.8 and 1.0 for PASCAL-Part, spanning different categories of objects (animals, tools and vehicles), and up to 14 visible parts per class. The F1-score depends on the quality of the ground truth, but also on the number of meronyms that compose each object. Increasing the quality of scraping, and thus the meronym model, could allow HOLMES to recognize and include in the provided explanations an even larger pool of meronyms.  In addition, HOLMES provides intrinsic safeguards against this type of noise, as only meronyms with sufficient F1-score are included in the explanations, and the user can readily inspect the heatmaps associated to each individual meronym. 

\LM{Quantitative causal metrics based on the deletion/insertion/preservation curves confirm that the part-based explanations provided by HOLMES are effective in identifying those parts that are most relevant to the final classification, achieving results comparable to the state-of-the-art GradCAM method, and substantially above chance level. However, unlike GradCAM, HOLMES provides an articulated set of heatmaps, associated to  human-interpretable concepts, and allows exploration of the impact of individual meronyms on the holonym classification, at both instance and class level. }

\LM{HOLMES was evaluated on two distinct datasets, PASCAL-Part and ImageNet. Since both pipelines evaluate the same classifier, we attribute absolute differences in the insertion/deletion/preservation curves to the dataset themselves (e.g., how the images were sourced), and possibly to the domain shift between ImageNet and PASCAL-Part (given that the holonym classifier $\mathcal{H}$ was trained on ImageNet). However, the relative performance with respect to both baselines shows similar behavior, despite wide differences in how the meronym datasets $X^c$ were sourced. HOLMES performs slightly better on PASCAL-Part, especially in terms of the deletion and preservation curves. Also, explanations on PASCAL-Part appear to be concentrated, on average, on fewer parts than ImageNet. Beyond the meronym datasets $X^c$, other factors could account for these differences: on the one hand, PASCAL-Part includes fewer and more distinct classes than ImageNet, thus potentially it includes images that are ‘easier’ to classify. On the other hand, the KB derived from PASCAL-Part annotations is simpler, with fewer meronyms ($\approx 4$ vs. $\approx 7$ parts per class), and less visual overlap. Overall, HOLMES shows to be robust to the choice of experimental settings, and performs well even when exploiting more cost-effective annotations sourced through general purpose KBs and web scraping.  }

\section{Conclusions and Future Work}
\label{sec:conclusions}

In this paper we introduced HOLMES, an eXplainable Artificial Intelligence technique able to enrich image classification tasks with part-level explanations.
Our approach allows to take a further step with respect to the standard label-level heatmaps which represent the state of the art in XAI for image classification. 
It proves valuable in integrating image classification models into decision support systems, as it provides more detailed explanations. These explanations can help both the model developer, aiding in debugging the classifier before deployment, and also the end user, assisting in assessing the level of trust in the classifier's predictions for previously unseen data.

Furthermore, HOLMES sheds light on how holonyms are learned and stored within a CNN during and after the training phase. Other recent research works proposed relevant contributions, such as~\cite{Bau_2020,Rudin}, but with additional requirements (focus on scene recognition, need for a segmented ground truth) and without the connection of a DL model with a symbolic knowledge base such as an ontology. 

We adopt a strategy that avoids confining concepts to a single computational unit, or neuron. We contend that this approach aligns better with the robust learning capabilities of DL models while also facilitating greater expressive power in the symbolic domain.
As shown in \cite{bau2017network}, models with equivalent discriminative abilities can exhibit varying degrees of interpretability in their units, which can be influenced by factors such as the model architecture, regularization and learning task.

Given the novelty of our proposed pipeline, there is room for exploration concerning alternatives for many components. First, a more refined scraping method could be employed to both increase the training sample size and its quality, for instance using more complex semantic expansion techniques, by using more robust outlier detection algorithms such as Robust and Kernel PCA, or by incorporating novel data purification techniques to obtain cleaner data \cite{zhang2021robust}. 

Second, it could be useful to study the effects of using the activations of units belonging to different convolutional layers, or even sets of such layers, for producing HOLMES explanations; units belonging to different convolutional layers could better match some specific (part) concepts, and, accordingly, by means of their activations, a better explanation could be hence generated for those concepts.
Also, more model architectures can be tested to see how this method results change according to the model which is used. For instance, shallower (e.g., VGG13) or deeper (e.g., VGG19) model architectures, or even different types of networks (e.g., Deep Residual Networks) could be inspected.

Third, alternative perturbation techniques can be tried for removing the relevant pixels of the parts; it was observed that substituting pixels with just constant values introduces contiguous shapes in the image, thus biasing, if even minimally, the prediction towards certain types of objects having a similar shape. Moreover, other types of semantic relationship can be studied for both retrieving the desired (visible) parts of a specific concept (either alternative or complementary to the proposed holonym-meronym relationship), and for mapping different concepts between different knowledge bases (in alternative to the proposed hypernym-hyponym relationship).

Finally, it emerges from the final results that the method performs better when considering for an object a small number of parts, preferably spaced enough to minimize visual overlap. Hence, a new strategy for selecting and filtering the meronyms of an object can be also studied.

\bibliographystyle{abbrv}
\bibliography{holmes_xai}

\clearpage
\section*{Appendix}
\label{sec:appendix}
In this Section, we show additional information and examples of part-based explanations generated by HOLMES.

\subsection{Holonym-Meronym PASCAL-Part mappings}
\begin{itemize}
\small
\item $ P_{bird} = P_{cat} = P_{dog} = P_{horse} = $ \{head, torso, leg, tail\}
\item $ P_{cow} = $ \{head, torso, leg, horn\}
\item $ P_{sheep} = $ \{head, torso, leg\}
\item $ P_{aeroplane} = $ \{stern, wheel, artifact wing, body, engine\}
\item $ P_{bicycle} = $ \{saddle, wheel, handlebar\}
\item $ P_{motorbike} = $ \{saddle, wheel, handlebar, headlight\}
\item $ P_{car} = P_{bus} = $ \{window, wheel, headlight, mirror, door, bodywork, license plate\}
\item $ P_{train} = $ \{coach, locomotive, headlight\}
\item $ P_{bottle} = $ \{body, cap\}
\end{itemize}

\subsection{Holonym-Meronym ImageNet mappings}

\begin{itemize}
\small
\item $ P_{sorrel} =  P_{zebra} =P_{ibex} = $ \{tail, legs, fur, hooves, feet, head\}
\item $ P_{hog} = $ \{tail, legs, feet, hooves, fur, curly tail, udder, head\}
\item $ P_{ox} = $ \{tail, legs, feet, hooves, fur, dewlap, head\}
\item $ P_{water buffalo} =P_{bison} $ \{hump, tail, legs, feet, hooves, fur,  head\}
\item $ P_{ram} =  $ \{legs, fur, hooves, feet, head\}
\item $ P_{arabian Camel} = $ \{long neck, tail, legs, feet, hooves, fur, hump, head\}
\item $ P_{tabby} = P_{tiger cat} = P_{persian cat} = P_{siamese cat} = P_{egyptian cat} = $ \{head, tail, legs, fur, feet \}
\item $ P_{cougar} = P_{lynx} = P_{leopard} = P_{snow leopard} = P_{jaguar} = P_{lion} = P_{cheetah} =$ \{head, tail, legs, fur, feet \}
\item $ P_{tiger} = $ \{head, tail, legs, fur, feet, ruff \}
\item $ P_{loggerhead} =  P_{leatherbackturtle} = P_{mudturtle} =  P_{terrapin} =P_{box turtle} =$ \{scales, shell, tail, head, flippers \}
\item $ P_{commoniguana} = $ \{scales, legs, feet, head, dewlap, tail \}
\item $ P_{americanalligator} = $ \{scales, legs, feet, head, tail \}
\item $ P_{rockpython} = $ \{scales, head \}
\item $ P_{diamondback} =P_{sidewinder} = $ \{scales, tail, head \}
\item $ P_{dishwasher} = $ \{racks, push-buttons, knob, door \}
\item $ P_{dutchover} =P_{rotisserie} = $ \{racks, knob, door \}
\item $ P_{microwave} = $ \{push-buttons, monitor, door \}
\item $ P_{stove} = $ \{knob, door, oven, racks, hob on top \}
\item $ P_{toaster} = $ \{push-buttons, knob, slots, lever \}
\item $ P_{broom} = $ \{handle, bristles \}
\item $ P_{hammer} = $ \{handle, head \}
\item $ P_{hatchet} = $ \{handle, head \}
\item $ P_{power drill} = $ \{ trigger,handle, threads, enclosure \}
\item $ P_{screw} = $ \{flat head, threads\}
\item $ P_{screwdriver} = $ \{handle, shaft, contoured head\}
\item $ P_{shovel} = $ \{handle, shaft, blade\}
\item $ P_{airliner} = $ \{nose, windows, wheels, turbines, door, wings, fins, propeller, tail \}
\item $ P_{ambulance} = $ \{coachwork, bumper, siren, windows, cargo space, wheels, door, number plate, mirror, bonnet, light, roof \}
\item $ P_{moving van} = $ \{coachwork, bumper, windows, cargo space, wheels, door, number plate, mirror, bonnet, light, roof \}
\item $ P_{barrow} = $ \{handle, draw bar, wheels, container\}
\item $ P_{beachwagon} = $ \{wheels, steering wheels, windows, mirror, bonnet, light, roof,number plate, bumper, door, trunk \}
\item $ P_{bicycle-built-for-two} = $ \{step-through frame, fork, wheels, chain, geras, handlebar, brakes, seat, pedals \}
\item $ P_{cab} =  P_{limousine} =$ \{coachwork, wheels, windows, mirror, bonnet, light, roof,number plate, bumper, door, trunk \}
\item $ P_{convertible} =  P_{freightcar} = P_{minivan} = P_{Model T}= P_{passengercar}=    P_{policevan} =  P_{racer} =  P_{sportscar} = $ \{wheels, steering wheels, windows, mirror, bonnet, light, roof,number plate, bumper, door, trunk, belts \}
\item $ P_{fireengine} =  P_{tailertruck} =  P_{pickup} = P_{towtruck} = $ \{coachwork, wheels, windows, mirror, bonnet, light, roof,number plate, bumper, door,  cabin, container, trailer \}
\item $ P_{horsecart} = P_{jinrikisha} =  $ \{handle, handlebar, wheels, container\}
\item $ P_{oxcart} =  P_{shoppingcart} =$ \{frame, handlebar, wheels, container\}
\item $ P_{jeep} =  P_{garbagetruck} = $ \{coachwork, wheels, windows, steering wheel, mirror, bonnet, light, roof,number plate, bumper, door,  trunk, spare tire \}
\item $ P_{minibus} =   P_{schoolbus} =  P_{trolleybus} = $ \{coachwork, wheels, windows, mirror,  light, roof,number plate, bumper, door\}
\item $ P_{mobilehome} =  $ \{windows, roof, wheels, draw bar, door\}
\item $ P_{moped} =  $ \{frame, seat, wheels, light, exhaust pipe, stand, mudguard, handlebar, tank, windshield, mirror\}
\item $ P_{motorscooter} =  $ \{step-through frame, seat, wheels, light, exhaust pipe, stand, mudguard, handlebar, floorboard, mirror\}
\item $ P_{mountainbike} =  $ \{step-through frame, seat, wheels, fork, chain, gears, handlebar, breaks, pedals\}
\item $ P_{streetcar} =   $ \{handlebar, wheels, platform\}
\item $P_{tractor} = $ \{coachwork, bumper, wheels, steering wheel,  bonnet, light, brakes,loader seat \}
\item $ P_{unicycle} =  $ \{frame, seat, wheel, fork, pedals\}
\end{itemize}

\subsection{Examples of part-based explanations}

\begin{figure*}[h]
\centering
\includegraphics[width=0.95\textwidth]{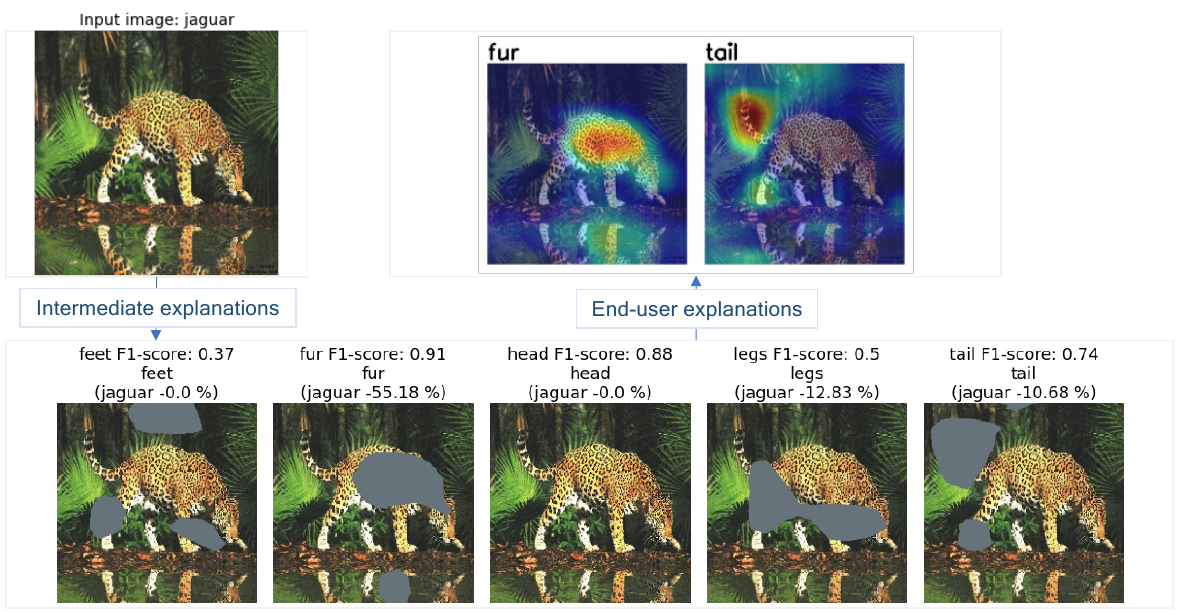}
\caption[Example n. 1.]{Example n. 1. The image portions associated to the meronyms (feet, fur, head, legs and tail) of the predicted holonym class (jaguar) are identified in the input image and ablated one at time. The feet and the legs calibrated F1-scores are below the calibrated F1-score threshold (0.7) and so these meronyms are immediately discarded: in fact, for such meronyms, the highlighted regions do not precisely encircle the respective part, while also including extraneous regions like background (vegetation) and other parts (the head and the tail are partially masked in the legs case). Among the remaining meronyms (fur, head and tail), only the fur and the tail score drops are above the holonym score drop threshold (10), hence only for these meronyms (in red) the associated saliency maps are interpolated with the input image and provided as End-user explanations. Notice as for this specific `jaguar' input image the head is not recognized: its associated image portion is not highlighted, hence a zero holonym score drop is recorded and consequently the part in not included in the End-user explanations.
}
\label{fig:jaguar_exp}
\end{figure*}

\begin{figure*}[h]
\centering
\includegraphics[width=0.97\textwidth]{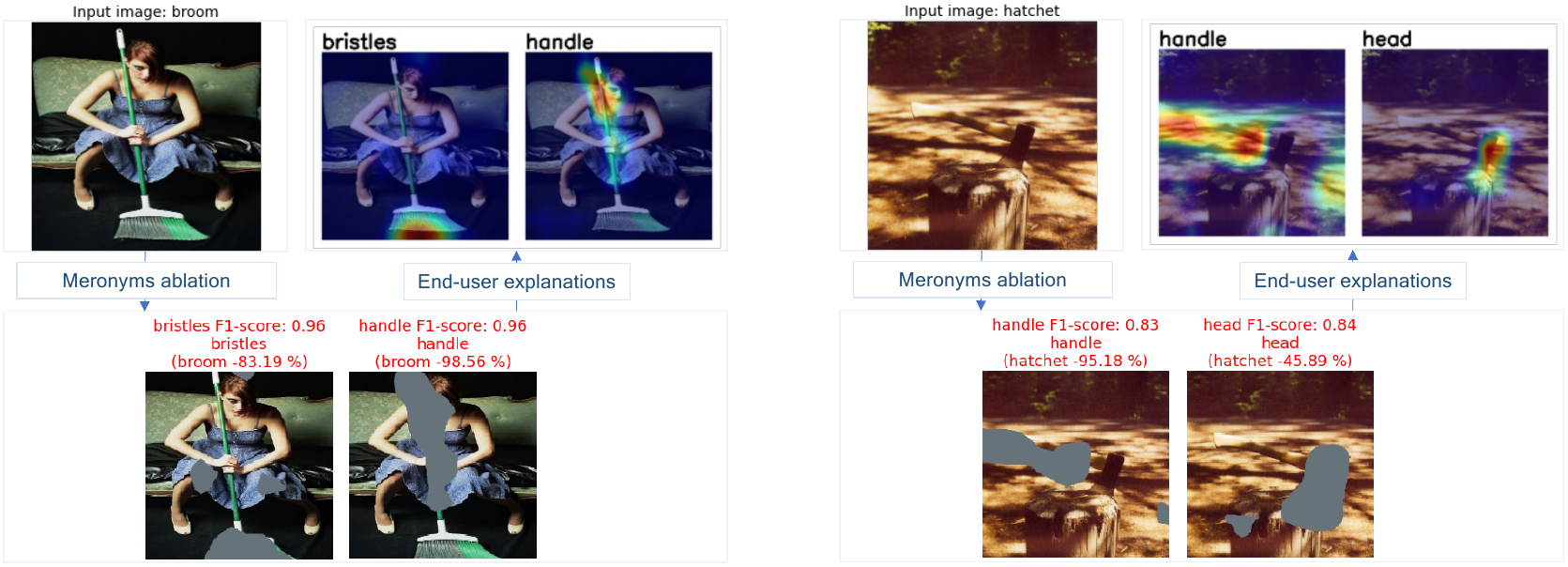}
\caption[Example n. 2.]{Example n. 2. Two different correctly predicted input images (broom on the left and hatchet on the right) get their respective parts identified and ablated. Since for all their meronyms both the calibrated F1-score threshold and the holonym score drop threshold is exceeded, all the parts (in red) are included in the End-user explanations. Notice as in both cases all the meronyms calibrated F1-scores are quite high (0.83 to 0.96), hence the image portions which are highlighted effectively describe the meronyms.
}
\label{fig:tools_exp}
\end{figure*}

\begin{figure*}[h]
\centering
\includegraphics[width=0.86\textwidth]{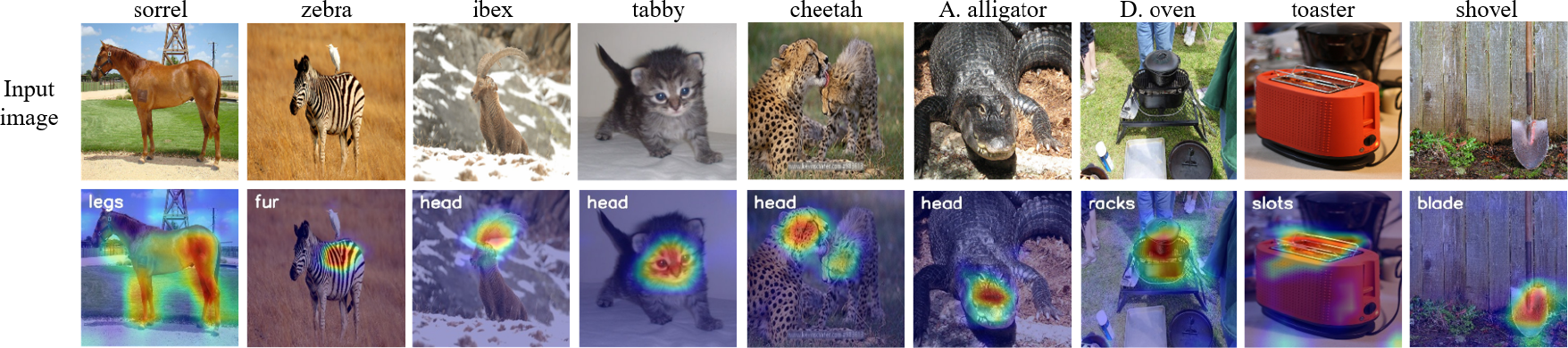}
\caption[batch1]{Examples of End-user explanations comprising 1 meronym.
}
\label{fig:batch1}
\end{figure*}

\begin{figure*}[h]
\centering
\includegraphics[width=0.86\textwidth]{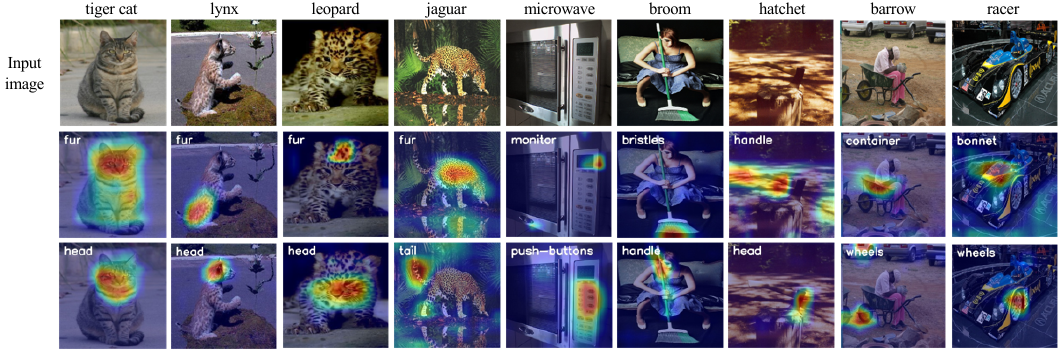}
\caption[batch2]{Examples of End-user explanations comprising 2 meronyms.
}
\label{fig:batch2}
\end{figure*}

\begin{figure*}[h]
\centering
\includegraphics[width=0.86\textwidth]{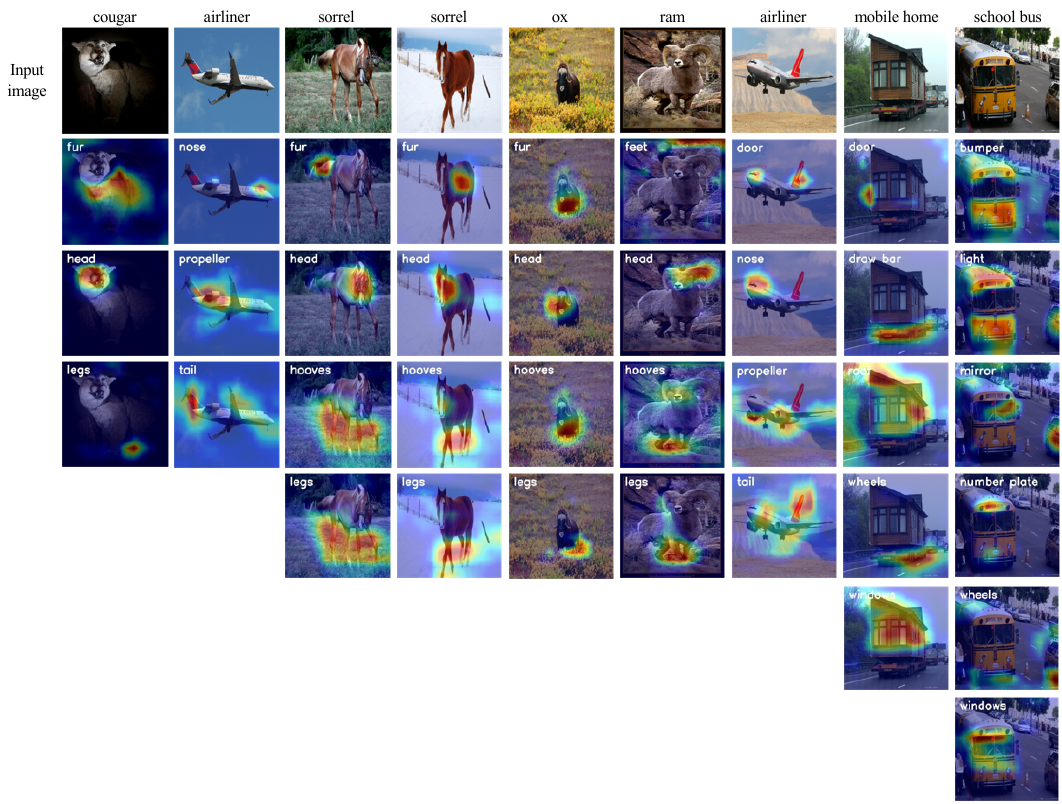}
\caption[batch3]{Examples of End-user explanations comprising 3 or more meronyms.
}
\label{fig:batch3}
\end{figure*}

Figures \ref{fig:jaguar_exp}, \ref{fig:tools_exp}, \ref{fig:batch1}, \ref{fig:batch2} and \ref{fig:batch3} show examples of part-based explanations extracted by HOLMES.


\end{document}